\documentclass{article}

\usepackage[numbers]{natbib}
\usepackage[preprint]{neurips_2026}

\usepackage[utf8]{inputenc} 
\usepackage[T1]{fontenc}    
\usepackage[table]{xcolor}  
\usepackage{hyperref}       
\usepackage{url}            
\usepackage{booktabs}       
\usepackage{amsfonts}       
\usepackage{nicefrac}       
\usepackage{microtype}      

\usepackage{graphicx}
\usepackage{amsmath}
\usepackage{enumitem}
\usepackage[most]{tcolorbox}
\usepackage{multirow}
\usepackage{subcaption}
\usepackage{animate}

\usepackage{algorithm}
\usepackage{algpseudocode}

\title{SWEET: Sparse World Modeling with Image Editing for Embodied Task Execution}

\author{
  Yiren Song$^{1}$\thanks{Equal contribution.} \quad
  Yihan Wang$^{1}$\footnotemark[1] \quad
  Xiyao Deng$^{1}$ \quad
  Zhuoran Yan$^{2}$ \quad
  Mike Zheng Shou$^{1}$\thanks{Corresponding author.} \\
  \\
  $^{1}$Show Lab, National University of Singapore \\
  $^{2}$Central South University
}

\begin{document}

\maketitle

\vspace{-7mm}

\begin{figure}[htbp]
\centering
\includegraphics[width=1.0\textwidth]{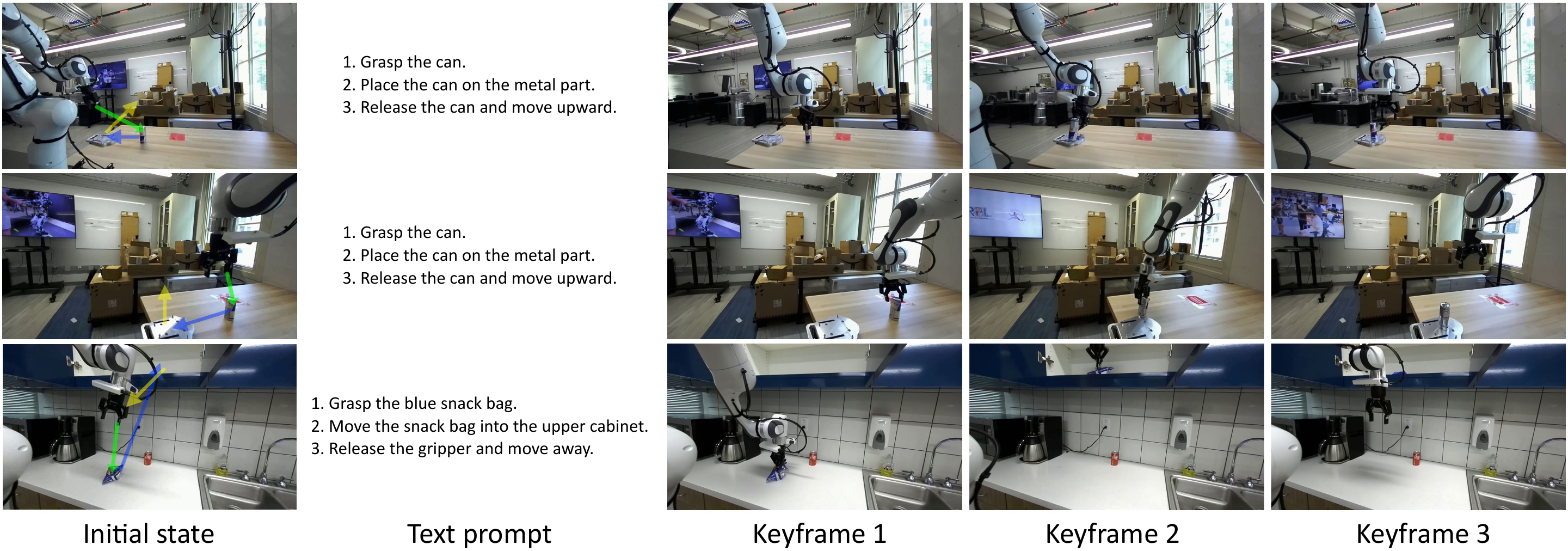}
\vspace{-4mm}
\caption{SWEET converts language-guided manipulation instructions into sparse visual keyframes through successive image editing, which are then executed by a goal-conditioned action predictor.}
\label{fig:teaser}
\end{figure}

\title{SWEET: Sparse World Editing for Embodied Task Execution}

\begin{abstract}
Visual prediction has emerged as a promising paradigm for embodied control, where future observations are generated and then translated into actions.
However, dense video generation is computationally expensive and often unnecessary for many manipulation tasks, whose progress can be summarized by a small number of task-relevant visual states.
In this work, we study whether image editing models can serve as sparse visual world models for robot manipulation by predicting task-level future states without dense video rollout.
We first conduct a controlled comparison between the video generation model Wan2.2 and the image editing model FLUX-Kontext under the same robotic data setting, and find that image editing produces more reliable task-level keyframes with better visual fidelity and substantially lower inference cost.
Motivated by this observation, we propose SWEET, a one-shot sparse visual planning framework that progressively generates a sequence of task-relevant manipulation keyframes through successive image editing, conditioned on language instructions and optional arrow-based spatial guidance.
A goal-conditioned diffusion action predictor then converts adjacent imagined keyframes into executable action chunks.
To reduce the mismatch between real and edited visual subgoals, we further introduce a mixed-training strategy with filtered edited targets.
Experiments on DROID and RoboMimic show that SWEET improves keyframe prediction across seen and unseen scenes and enables a full pipeline from sequential keyframe planning to executable robot actions, suggesting that image editing is a promising and underexplored direction for embodied visual prediction. Code is released at \href{https://github.com/showlab/SWEET}{https://github.com/showlab/SWEET}
\end{abstract}

\section{Introduction}

Recent advances in video generation have made visual prediction an important foundation for embodied intelligence.
By imagining future observations, video models can serve as generative world models or world-action models for planning, inverse dynamics, and robot control.
However, many robotic manipulation tasks can be specified by a small number of task-relevant keyframes, such as approach, contact, grasp, transport, and release, raising a natural question: \emph{is dense video prediction truly necessary for robotic manipulation?}
This motivates us to study sparse keyframe prediction as a more efficient alternative to dense video rollout.

Generated visual goals have started to emerge as useful intermediate representations for robot control, including image-based subgoal synthesis~\cite{black2023zero}, goal-image-conditioned policy learning~\cite{li2025gr}, and object-centric goal-state generation~\cite{chen2025goal}.
These studies suggest that visual future states can provide an explicit and interpretable interface between task intent and robot execution.
However, existing efforts typically use generated images as single-step goals, auxiliary policy conditions, or final-state representations.
It remains underexplored whether modern in-context image-editing models can serve as sparse visual world models that successively imagine a sequence of task-relevant future states, especially in comparison with dense video generation under the same robotic data setting.

To answer this question, we conduct a controlled comparison between video generation and image editing for robotic keyframe prediction.
We construct train/test splits on DROID and fine-tune representative open-source backbones, including \textbf{Wan2.2} and \textbf{FLUX Kontext}, under the same task conditions.
Our results show that image editing produces more reliable task-level keyframes than dense video generation, with better visual fidelity, scene consistency, robot-arm preservation, and inference efficiency. This suggests that modern image-editing models can provide a promising sparse keyframe-level planning prior for robotic manipulation.

Motivated by this finding, we propose \textbf{SWEET}, a sparse world-modeling framework with image editing for embodied task execution.
SWEET uses an image-editing model to progressively generate a sequence of future keyframes from the initial observation.
The generation condition can be specified by natural language instructions, optional arrow-based spatial guidance, or their combination.
The arrow provides a coarse high-level spatial hint, which can be specified by a user or predicted by a vision-language or vision-reasoning model.
During inference, SWEET renders the corresponding arrow cue on the current keyframe and applies image editing to produce the next visual subgoal.
By repeating this process, it constructs a compact sequence of task-relevant keyframes without requiring dense video rollouts.

To connect sparse visual imagination with robot execution, we further train a goal-conditioned diffusion action predictor to infer executable action chunks between adjacent imagined keyframes.
Since the predictor is trained on clean keyframe transitions but receives editing-generated subgoals at inference time, a real-to-edited domain gap naturally arises.
To reduce this mismatch, we introduce a direct mixed-training strategy, where edited keyframes are generated by the visual world model, filtered for quality, and mixed with clean transitions during action-predictor training.
This exposes the action predictor to the visual distribution of generated targets and improves robustness during execution.

Evaluations on both the DROID dataset and RoboMimic simulation support our hypothesis.
On DROID, SWEET improves real-world keyframe prediction across seen and unseen scenes.
On RoboMimic, full-pipeline simulation experiments show that the generated visual subgoals can be translated into executable robot actions under limited training data.
Overall, our findings suggest that image editing models are an underexplored yet promising foundation for embodied visual prediction, enabling a complete pipeline from sequential keyframe imagination to executable robot action.

Our contributions are summarized as follows:
\begin{itemize}
    \item We compare dense video generation and image editing under the same robotic data setting, revealing image editing as a promising prior for sparse keyframe-level future-state modeling.

    \item We propose \textbf{SWEET}, a sparse world-modeling framework that successively edits images into task-relevant future keyframes conditioned on language and optional arrow guidance.
    
    \item We reduce the real-to-edited subgoal gap with mixed training on filtered edited targets, and validate SWEET on DROID keyframe prediction and RoboMimic full-pipeline execution.
\end{itemize}

\section{Related Work}

\subsection{Vision Language Action Models.}
Vision-Language-Action (VLA) models represent a major direction for building foundation models in robotics. 
Early approaches use pretrained language or vision-language models as high-level planners that generate instructions, programs, or visual reasoning traces, which are then grounded by separate low-level robotic controllers~\citep{ahn2022can,huang2022inner,liang2023code,huang2023voxposer,driess2023palm}. 
This modular design improves semantic reasoning and generalization, but often depends on predefined skills, perception APIs, or carefully designed planning-execution interfaces. 
In parallel, language-conditioned imitation learning methods train visuomotor policies directly from demonstrations, using language or multimodal prompts to specify manipulation goals~\citep{shridhar2022cliport,jang2022bc,mees2022calvin,mees2022matters,jiang2023vima}. End-to-end VLAs integrate language understanding, visual perception, and low-level actions within a single model. 
Representative systems include generalist agents and robotic transformers trained on large-scale robot datasets or mixed-embodiment data~\citep{reed2022generalist,brohan2022rt,zitkovich2023rt,bousmalis2023robocat,collaboration2023open,team2024octo,kim2024openvla,black2024pi0}. 
While these models show strong semantic grounding and object-level generalization, they often rely on large teleoperation datasets or web-scale vision-language pretraining, and do not explicitly model future task states.

\subsection{Video Model-based Robot Policies.}
Video generation models have been increasingly used to support robot policy learning \cite{yuan2026fast,  liang2025video, zhen2026action}. 
Early visual foresight methods use learned video prediction models for model-predictive control from pixels~\citep{finn2017deep}, while recent generative-policy methods cast decision making as text-guided video generation followed by inverse dynamics, goal-conditioned control, or hierarchical planning~\citep{du2023learning,du2023video, ajay2023compositional, song2025mitty, yang2025x, song2026omnihumanoid}. 
Other works leverage large-scale video pretraining or compositional video world models to synthesize robot behaviors, generate future plans, and improve generalization to unseen tasks and environments~\citep{wu2023unleashing,wan2025wan}. 
These approaches show that video models can provide rich visual dynamics priors for robot learning. Another line of work jointly models future visual states and robot actions, or uses generated future observations as subgoals for downstream controllers~\citep{chi2025diffusion,janner2022planning,ajay2022conditional,shafiullah2022behavior,florence2022implicit}. 
These methods are closely related to World Action Models, where world prediction and action prediction are coupled for decision making. 
However, dense video rollout is computationally expensive and often contains redundant intermediate frames for manipulation tasks that can be summarized by sparse semantic milestones. 
Our work therefore studies a complementary formulation: instead of generating full future videos, SWEET directly predicts task-relevant keyframes through image editing and uses them as compact visual subgoals for action prediction.

\subsection{Image Editing Models.}
Recent image editing models have moved from prompt-only synthesis toward image-conditioned  and in-context visual transformation, where an input image is modified according to textual or visual instructions~\citep{brooks2023instructpix2pix,zhang2023magicbrush,sheynin2024emu,labs2025flux,xiao2025omnigen, liu2025omnirefiner, song2025omniconsistency, song2025makeanything, huang2025arteditor, gong2025relationadapter, wang2025diffdecompose, yan2025eedit}. 
Unlike pure text-to-image generation, these models are trained on paired transformation data, including manually annotated edits, synthetic instruction-edit pairs, and temporally related image pairs sampled from videos \cite{wu2025qwen, batifol2025flux}. 
Such supervision encourages the model to preserve the source scene while applying semantically meaningful changes, making it naturally aligned with robotic keyframe prediction: the background, object identity, and scene layout should remain stable, while the robot and task-relevant objects evolve according to language or spatial guidance. 
Recent work such as SuSIE shows that pretrained image-editing diffusion models can propose visual subgoals for robotic manipulation~\citep{black2023zero}. 
Our work follows this direction but focuses on fine-tuned keyframe-based visual planning with explicit spatial guidance and a diffusion-based action predictor trained to handle both real and edited subgoal domains.

\section{Methods}

We now describe the SWEET framework in detail.
We first formulate sparse keyframe-based visual planning for manipulation, then introduce how keyframe supervision is constructed from robot trajectories.
We next present the image-editing planner, the goal-conditioned action predictor, and the mixed-training strategy for reducing the real-to-edited subgoal gap, followed by the inference procedure.

\begin{figure*}[tbp]
    \centering
    \includegraphics[width=\textwidth]{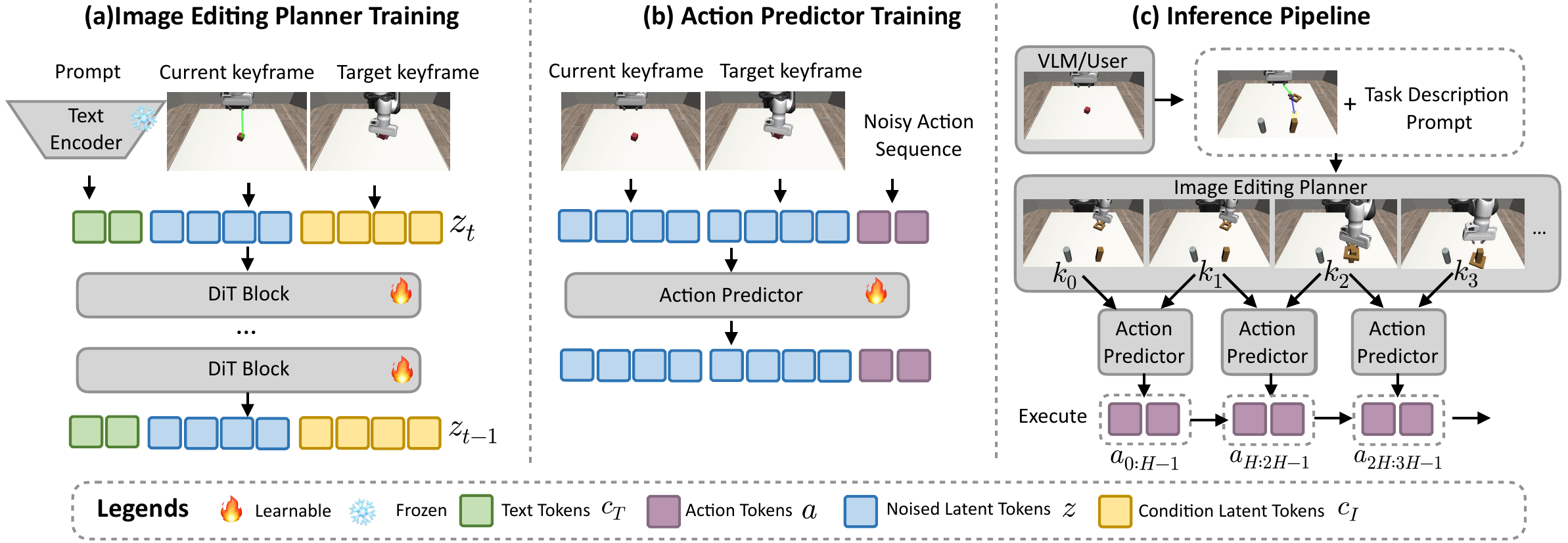}
    \vspace{-5mm}
    \caption{Overview of SWEET. 
    SWEET first trains an image editing planner to imagine task-relevant future keyframes, and then trains a goal-conditioned action predictor to execute actions between adjacent visual subgoals. 
    At inference time, the planner generates a sparse keyframe plan from the initial observation, and the action predictor translates each planned keyframe transition into executable action chunks.}
    \label{fig:method_overview}
\end{figure*}

\subsection{Problem Formulation}

We consider an embodied manipulation trajectory 
\(\tau=\{(x_t,a_t)\}_{t=1}^{T}\), where \(x_t\) denotes the RGB observation and \(a_t\) denotes the robot action at time step \(t\).
Each trajectory is represented by a sparse sequence of task-relevant keyframes 
\(\mathcal{K}=\{k_0,k_1,\ldots,k_M\}\), where \(k_m=x_{t_m}\) and 
\(t_0<t_1<\cdots<t_M\).
These keyframes summarize the main semantic stages of a manipulation task, such as approach, contact, grasp, transport, and release.
The action sequence between two neighboring keyframes is denoted as
\[
A_m=(a_{t_m},a_{t_m+1},\ldots,a_{t_{m+1}-1}).
\]

For each keyframe transition \(k_m\rightarrow k_{m+1}\), we define a subtask condition 
\(c_m=(l_m,r_m)\), where \(l_m\) is a language instruction and \(r_m\) is an optional arrow-based spatial cue.
At inference time, the input to SWEET consists of an initial observation \(o_0\) and a sequence of subtask conditions 
\(\mathcal{C}=\{c_m\}_{m=0}^{M-1}\).
The goal is to predict a sparse visual plan 
\(\hat{\mathcal{K}}=\{\hat{k}_0,\hat{k}_1,\ldots,\hat{k}_M\}\), with \(\hat{k}_0=o_0\), and then infer executable action chunks that move the robot through these planned visual subgoals:
\[
\hat{\mathcal{K}} = E_\theta(o_0,\mathcal{C}),
\qquad
\hat{A}_{t:t+H-1}=\pi_\phi(o_t,\hat{k}_{m+1}).
\]
Here, \(E_\theta\) denotes the recursive application of the image-editing planner over the subtask conditions, \(\pi_\phi\) denotes the goal-conditioned action predictor, and \(H\) is the action horizon.
Thus, the problem is formulated as sparse visual planning followed by goal-conditioned action execution, rather than dense future video prediction.

\subsection{Keyframe Dataset Construction}

We construct keyframe-based supervision from trajectory videos with paired visual observations and robot actions. 
For each episode, we manually annotate a sparse set of task-critical keyframes, corresponding to stages such as approaching the object, making contact, grasping, moving toward the target, placing, or releasing. 
For a trajectory, the annotation yields a keyframe sequence \(\mathcal{K}=\{k_0,k_1,\ldots,k_M\}\), where each neighboring pair \(k_m\rightarrow k_{m+1}\) defines a subtask-level transition. 
The low-level actions between the two timestamps are used as the action supervision for this transition.

For each keyframe transition, we further annotate a subtask-level text description and a lightweight 2D spatial cue. 
The text description summarizes the semantic change between two keyframes, such as moving toward the object, closing the gripper, or placing the object at the target location. 
The spatial cue is rendered on the source keyframe as an arrow-overlaid image \(k^{\mathrm{arr}}_m\), indicating the coarse image-space motion direction of the gripper. 
Its color encodes the gripper-state change, such as open-to-close or close-to-open, and a circle is used instead of an arrow when the gripper mainly acts in place with negligible displacement. 
Overall, each annotated transition contains \((k_m,k^{\mathrm{arr}}_m,l_m,k_{m+1},A_m)\). 
We use \((k^{\mathrm{arr}}_m,l_m,k_{m+1})\) to supervise the image-editing planner, while \(A_m\) is used for training the subgoal-conditioned action predictor. 

For the training set and benchmarks in this work, both keyframes and spatial cues are manually annotated to ensure reliable supervision. 
We also tested VLM-based annotation for the arrow cues and found that models such as Gemini-3 can reasonably localize gripper motion and produce arrow annotations when provided with sufficient contextual information. 
In contrast, extracting task-critical keyframes from videos, especially contact and state-transition moments, remains unreliable with current video understanding models. 
Therefore, we adopt human annotation for keyframe selection in this work.

\subsection{Image Editing Planner}
\vspace{-2mm}

We instantiate the visual planner with a pretrained image-editing model.
In our main experiments, we build the planner on \textbf{FLUX Kontext} and adapt it to robotic manipulation through LoRA fine-tuning.
Given the current keyframe \(k_m\), the subtask text description \(l_m\), and an optional arrow-based spatial cue \(r_m\), the planner predicts the next task-relevant keyframe:
\vspace{-0.2em}
\begin{equation}
\hat{k}_{m+1}=E_\theta(k_m,l_m,r_m).
\label{eq:image_editing_planner}
\end{equation}
\vspace{-0.2em}
During training, each sample is constructed from an annotated transition \((k_m,l_m,r_m,k_{m+1})\), where \(k_{m+1}\) is the target keyframe.
The text description specifies the subtask-level semantic change, while the optional arrow provides coarse spatial guidance when available.
To support both language-only and language-plus-arrow planning, we randomly drop the arrow condition during training while always keeping the text description.
We optimize the planner with the native image-editing objective of FLUX Kontext and update only the inserted LoRA parameters.

At inference time, SWEET starts from the initial observation \(\hat{k}_0=k_0\) and successively generates future keyframes through repeated image-editing steps, each conditioned on the corresponding subtask text and optional arrow guidance.
This yields a compact sequence of visual subgoals that captures the intended task progress without requiring dense video rollout.

\subsection{Goal-conditioned Diffusion Action Predictor}
\vspace{-2mm}

After generating the visual keyframe plan, SWEET executes each predicted keyframe as a subgoal.
To this end, we train a goal-conditioned diffusion action predictor \(\pi_\phi\).
Given the current observation \(o_t\) and a target keyframe \(\hat{k}_{m+1}\), the predictor outputs an action chunk:
\begingroup
\setlength{\abovedisplayskip}{3pt}
\setlength{\belowdisplayskip}{3pt}
\setlength{\abovedisplayshortskip}{3pt}
\setlength{\belowdisplayshortskip}{3pt}
\begin{equation}
\hat{A}_t=\pi_\phi(o_t,\hat{k}_{m+1}).
\label{eq:action_predictor}
\end{equation}
\endgroup
For training, we use keyframe intervals from the annotated trajectories.
Let
\begingroup
\setlength{\abovedisplayskip}{3pt}
\setlength{\belowdisplayskip}{3pt}
\setlength{\abovedisplayshortskip}{3pt}
\setlength{\belowdisplayshortskip}{3pt}
\begin{equation}
A_m=(a_{t_m},a_{t_m+1},\ldots,a_{t_{m+1}-1})
\label{eq:action_sequence}
\end{equation}
\endgroup
denote the action sequence between \(k_m\) and \(k_{m+1}\).
Since different keyframe intervals may have different lengths, we train the predictor in a receding-horizon manner with a fixed action horizon \(H=16\).
Each training target is an action chunk of length \(H\), and the corresponding conditioning input consists of the current observation and the target keyframe of the current subtask.

Formally, the action predictor is trained with the standard diffusion-policy denoising objective:
\begingroup
\setlength{\abovedisplayskip}{3pt}
\setlength{\belowdisplayskip}{3pt}
\setlength{\abovedisplayshortskip}{3pt}
\setlength{\belowdisplayshortskip}{3pt}
\begin{equation}
\phi^{*}
=
\arg\min_{\phi}
\mathcal{L}_{\mathrm{act}}(o_t,k_{m+1},A_t^{H}),
\label{eq:action_loss}
\end{equation}
\endgroup
where \(A_t^{H}=(a_t,\ldots,a_{t+H-1})\) is the fixed-horizon action chunk and \(\mathcal{L}_{\mathrm{act}}\) denotes the diffusion action-generation loss. During execution, the policy predicts an action chunk toward the current target keyframe, executes it, and then receives an updated observation.
If the target has not been reached, the predictor is called again with the updated observation and the same target keyframe.
Thus, although the high-level keyframe plan is generated open-loop, each subgoal is tracked in a closed-loop, receding-horizon manner.

\subsection{Mixed Training with Edited Transitions} 
\vspace{-2mm}

A practical challenge is the domain gap between training and inference for the action predictor. During standard training, the predictor observes clean transitions from the dataset, where the target subgoal is a real keyframe $k_{m+1}$. At inference time, however, the target subgoal is generated by the image-editing planner as $\hat{k}_{m+1}$. Even when the generated keyframe is semantically plausible, it may introduce subtle changes in texture, lighting, robot appearance, object boundaries, or background statistics, shifting the input distribution of the action predictor. To reduce this mismatch, we construct synthetic edited transitions. For each ground-truth transition, we use the visual planner to generate an edited target keyframe $\tilde{k}_{m+1}$, and pair it with the same action supervision from the original transition. This augmentation exposes the action predictor to the visual distribution of planner-generated subgoals, rather than creating new physical trajectories. Before training, we filter low-quality edited samples according to visual consistency and task plausibility, removing cases with obvious background distortion, object disappearance, severe robot inconsistency, or incorrect task state. We then combine these filtered edited transitions with the original clean transitions, and train the action predictor with a 1:1 mixture of both. Equivalently, each training batch contains equal numbers of clean and edited samples. This direct mixed-training strategy reduces the real-to-edited domain gap and improves robustness when the controller is deployed with planner-generated subgoals.

\subsection{Inference Procedure}
\vspace{-2mm}

At inference time, SWEET first generates a sparse keyframe plan and then executes the planned visual subgoals with the goal-conditioned action predictor.
Given an initial observation \(o_0\), we set \(\hat{k}_0=o_0\) and use the image-editing planner to successively generate a sequence of future keyframes before action execution.
For each subtask \(m\), the planner takes the current planned keyframe, the subtask text, and an optional arrow-based spatial cue as input:
{
\setlength{\abovedisplayskip}{3pt}
\setlength{\belowdisplayskip}{3pt}
\setlength{\abovedisplayshortskip}{2pt}
\setlength{\belowdisplayshortskip}{2pt}
\begin{equation}
\hat{k}_{m+1}=E_\theta(\hat{k}_m,l_m,r_m), \qquad m=0,\ldots,M-1.
\end{equation}
}
This produces a compact keyframe plan \(\hat{\mathcal{K}}=\{\hat{k}_0,\hat{k}_1,\ldots,\hat{k}_M\}\), where each adjacent pair \((\hat{k}_m,\hat{k}_{m+1})\) represents one task-relevant visual transition.

After obtaining the keyframe plan, SWEET executes the planned subgoals sequentially.
For the \(m\)-th transition, the next planned keyframe \(\hat{k}_{m+1}\) is used as the visual target.
Consistent with the training formulation of the action predictor, the policy takes the current observation \(o_t\) and the target keyframe \(\hat{k}_{m+1}\) as input, and predicts an executable action chunk:
{
\setlength{\abovedisplayskip}{3pt}
\setlength{\belowdisplayskip}{3pt}
\setlength{\abovedisplayshortskip}{2pt}
\setlength{\belowdisplayshortskip}{2pt}
\begin{equation}
\hat{A}_{t:t+H-1}=\pi_\phi(o_t,\hat{k}_{m+1}).
\end{equation}
}
The predicted action chunk is executed in the environment, after which the robot receives an updated observation.
The system then continues to track the next planned keyframe until all visual subgoals are executed.
Thus, in our current setting, the high-level keyframe plan is generated once, while low-level action execution is conditioned on online observations in a receding-horizon manner.

This one-shot sparse planning setup is used in our experiments because the evaluated tasks are short-horizon and can be described by only a small number of keyframes.
For more complex procedural or long-horizon tasks, the same planner--predictor interface can be naturally extended to a feedback-enabled setting.
After executing an action chunk and receiving a new observation \(o_t\), the planner can replan from the actual visual state and predict the next subgoal.
Our current evaluation focuses on the one-shot sparse planning setting, while feedback-based replanning is left as a natural extension.

\section{Experiments}
\vspace{-3mm}

\subsection{Experimental Setup}
\vspace{-2mm}

\paragraph{Experimental Scope.}
Our experiments assess image editing as a high-level visual planner for robotic manipulation, rather than proposing a state-of-the-art visuomotor policy. 
We focus on three aspects: sparse keyframe prediction against video generation, execution with a goal-conditioned action predictor, and mixed training for reducing the real-to-edited subgoal gap.

\paragraph{Implementation Details.}
Our framework consists of a high-level image-editing planner and a low-level goal-conditioned action predictor. 
For the planner, we use pretrained \textbf{FLUX.1-Kontext-dev}, including its DiT denoiser, text encoders, and autoencoder. 
We fine-tune only the DiT backbone with LoRA of rank 32 for 6K optimization steps on a single NVIDIA H20 GPU, using AdamW with a learning rate of \(1\times10^{-4}\). 
During training, the arrow-overlaid start frame of each subtask and the corresponding text prompt are used as the conditional editing input, while the subtask end frame is used as the target frame. 
At inference time, we set the editing horizon according to task complexity: in RoboMimic, the Lift task uses two editing steps corresponding to contact and lifting, while the Can and Square tasks use three editing steps; for DROID, all evaluated tasks use three editing steps.

For action prediction, we adopt the Diffusion Policy architecture and modify it into a start--goal conditioned policy that takes the current observation and target keyframe as visual conditions. The current observation and predicted subgoal image are encoded independently by a visual encoder, and their features are concatenated as the conditioning input to the action diffusion model. 
The policy predicts action chunks of length 16 and is trained with the standard DDPM denoising objective using AdamW, a learning rate of \(1\times10^{-4}\), and batch size 32. 
At inference time, it iteratively denoises a Gaussian action sequence conditioned on the current observation and subgoal, and executes the predicted chunk in a receding-horizon manner.

\begin{figure*}[tbp]
    \centering
    \includegraphics[width=\textwidth]{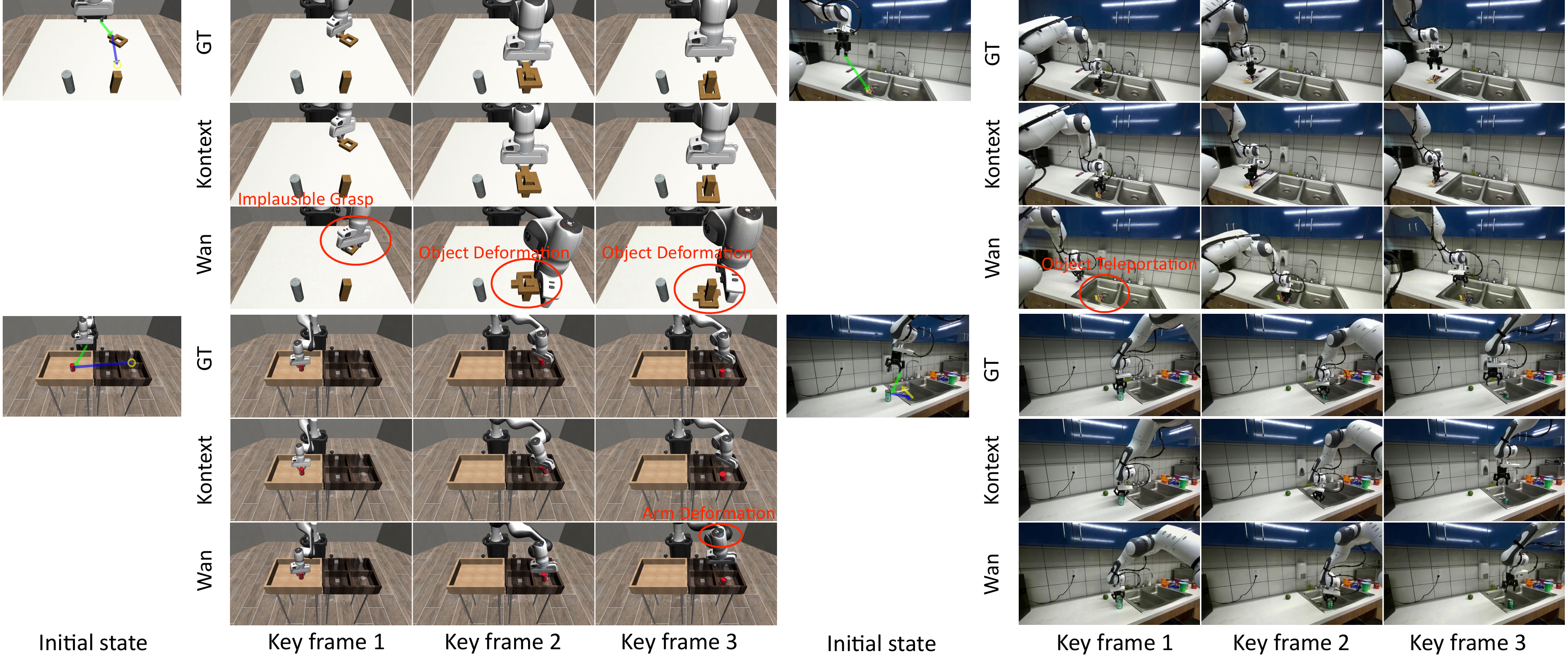}
    \vspace{-5mm}
    \caption{Qualitative comparison of keyframe planning on RoboMimic and DROID. 
    Compared with Wan2.2, the image-editing planner FLUX-Kontext generates more reliable task-relevant keyframes, with better embodiment consistency, more plausible object interactions, and fewer artifacts. 
    Wan2.2 more often suffers from missing or duplicated objects, distorted robot arms or grippers, inconsistent targets, and background mismatch, especially on the more complex DROID scenes.}
    \label{compare}
\end{figure*}

\paragraph{Datasets and Benchmarks.}
We evaluate SWEET on a real-robot DROID subset and the RoboMimic simulation benchmark. 
For DROID, we build a pick-and-place dataset with 700 samples and approximately 2,100 annotated subtasks across more than ten environments, covering variations in objects, backgrounds, layouts, and lighting. 
We use DROID to assess real-robot keyframe prediction under cross-scene generalization, with two 50-sample test sets for seen-scene and unseen-scene settings. 
For RoboMimic, we construct a simulation dataset with 600 samples and 1,600 subtasks over three manipulation tasks: Lift, Can, and Square. 
RoboMimic is used to evaluate both keyframe-generation quality and downstream execution in simulation. 
For each RoboMimic task, we run 200 evaluation trials to compute the task success rate.
We report image-level keyframe metrics on both benchmarks and task success rates on Lift, Can, and Square in RoboMimic.

\paragraph{Metrics.}
We evaluate keyframe quality using MSE, PSNR, SSIM, and LPIPS between generated and ground-truth keyframes.
For RoboMimic, we additionally report the environment-defined task success rate after deploying the full planner--predictor pipeline in simulation. We report endpoint MSE between executed and ground-truth end-effector positions for each subtask.

\subsection{Image Editing vs. Video Generation as Visual Planners}
\vspace{-2mm}

We first compare image editing and video generation as visual planners under the same robotic data setting. 
For the video-generation baseline, we use \textbf{Wan2.2} in its text-image-to-video (TI2V) setting. 
Given the arrow-overlaid initial frame and the subtask text description, Wan2.2 generates a dense future video. 
Since our benchmark provides annotated subtask stages, we select from the generated video the frame corresponding to the same stage as the target keyframe, such as contact, grasping, placement, or release, and use it for comparison with the ground-truth keyframe. 
For the image-editing planner, \textbf{FLUX-Kontext} directly predicts the next keyframe from the same arrow-overlaid input frame and text condition. 
This protocol evaluates both methods on the same sparse keyframe targets, while allowing the video model to produce a full rollout before keyframe selection.

\paragraph{Quantitative comparison.}
We evaluate predicted keyframes on RoboMimic and DROID using MSE, PSNR, SSIM, and LPIPS.
As shown in Table~\ref{tab:keyframe_metrics}, FLUX-Kontext consistently outperforms Wan2.2 across both simulation and real-robot datasets.
On RoboMimic, it achieves lower MSE and LPIPS as well as higher PSNR and SSIM, indicating better pixel-level fidelity, structural consistency, and perceptual similarity.
On DROID, it also performs better on both seen-scene and unseen-scene splits, showing stronger robustness to diverse backgrounds, object layouts, and camera viewpoints.
These results suggest that image editing provides a more reliable visual planning prior than dense video generation for sparse keyframe planning.

\paragraph{Qualitative analysis.}
Figure~\ref{compare} shows that the qualitative gap is more pronounced than the numerical metrics alone suggest.
FLUX-Kontext better preserves embodiment consistency, maintaining stable robot arms, grippers, and manipulated objects across sparse keyframes.
It also produces more plausible task states, especially in terms of grasping, object shape, robot-arm structure, and object location.
In contrast, Wan2.2 often exhibits four representative planning-level failures: implausible grasps, where the gripper does not form a physically valid contact with the object; object deformation, where the manipulated object changes its shape or identity; arm deformation, where the robot arm or gripper becomes structurally distorted; and object teleportation, where the object appears to move discontinuously across frames. These errors are critical for robotic planning, since predicted keyframes serve as target states for the controller rather than mere image-generation outputs.

\paragraph{Efficiency and discussion.}
Image editing also shows a clear inference-efficiency advantage in our evaluated setting. 
On a NVIDIA H20 GPU, FLUX-Kontext completes one task-level visual plan in approximately 10 seconds by generating 3 sparse keyframes, whereas Wan2.2 takes over 400 seconds to produce an 81-frame dense rollout. 
This efficiency gap reflects the difference between sparse keyframe prediction and dense video rollout: for many manipulation tasks, the planner only needs to predict task-relevant semantic milestones rather than all intermediate frames. 
Together with the reconstruction and perceptual results, this suggests that image editing offers a practical, lightweight visual-planning prior for robotic manipulation under current open-source models.


\begin{figure*}[tbp]
    \centering
    \animategraphics[width=\textwidth]{5}{pictures/fig4/frame_}{00}{09}
    \vspace{-5mm}
    \caption{Qualitative visualization of SWEET on the RoboMimic simulation benchmark, covering Lift, Can, and Square. The visualized pipeline includes keyframe planning, action prediction, and execution. Readers can click and play the video using {\color{red}\textbf{Adobe Acrobat}}.}
    \label{fig:closed_loop_rollout}
\end{figure*}

\begin{table}[tbp]
\centering
\caption{Controlled comparison of robotic keyframe prediction. Under the same data setting, FLUX-Kontext achieves better reconstruction and perceptual similarity to annotated keyframes than Wan2.2 on RoboMimic and DROID.}
\label{tab:keyframe_metrics}
\footnotesize
\setlength{\tabcolsep}{5pt}
\begin{tabular}{llcccc}
\toprule
Dataset & Method & MSE$\downarrow$ & PSNR$\uparrow$ & SSIM$\uparrow$ & LPIPS$\downarrow$ \\
\midrule
\multirow{2}{*}{RoboMimic}
& Wan 2.2 TI2V 5B & 0.01777 & 19.00 & 0.8563 & 0.1303 \\
& FLUX-Kontext    & \textbf{0.01114} & \textbf{20.71} & \textbf{0.9119} & \textbf{0.07832} \\
\midrule
\multirow{2}{*}{DROID-seen scene}
& Wan 2.2 TI2V 5B & 0.02180 & 17.35 & 0.8030 & 0.2032 \\
& FLUX-Kontext    & \textbf{0.01896} & \textbf{18.14} & \textbf{0.8366} & \textbf{0.1570} \\
\midrule
\multirow{2}{*}{DROID-unseen scene}
& Wan 2.2 TI2V 5B & 0.02844 & 15.64 & 0.7337 & 0.2625 \\
& FLUX-Kontext    & \textbf{0.02682} & \textbf{16.02} & \textbf{0.7608} & \textbf{0.2343} \\
\bottomrule
\end{tabular}
\end{table}

\begin{figure*}[t]
\centering

\begin{minipage}[t]{0.60\textwidth}
\centering
\captionof{table}{Action predictor ablation on RoboMimic. We report task success rate (\%) and action prediction MSE. Here, ``Generated keyframe'' means that the target keyframe is produced by the image-editing planner, while ``Real keyframe'' refers to the keyframe rendered by RoboMimic.}
\label{tab:action_predictor_ablation_all}
\vspace{-2mm}
\scriptsize
\setlength{\tabcolsep}{1.4pt}
\renewcommand{\arraystretch}{1.08}
\resizebox{\linewidth}{!}{
\begin{tabular}{llcc|cc|cc}
\toprule
\multirow{2}{*}{Task} 
& \multirow{2}{*}{Test target}
& \multicolumn{2}{c|}{Real-trained}
& \multicolumn{2}{c|}{Gen-trained}
& \multicolumn{2}{c}{Mix-trained} \\
\cmidrule(lr){3-4} \cmidrule(lr){5-6} \cmidrule(lr){7-8}
& & SR $\uparrow$ & MSE $\downarrow$
  & SR $\uparrow$ & MSE $\downarrow$
  & SR $\uparrow$ & MSE $\downarrow$ \\
\midrule

\multirow{2}{*}{Lift}
& Generated keyframe
& 58.0 & 0.0159
& 91.5 & 0.0171
& \textbf{92.0} & \textbf{0.0141} \\
& Real keyframe
& 87.0 & 0.0165
& 78.5 & \textbf{0.0144}
& \textbf{92.0} & 0.0145 \\

\midrule

\multirow{2}{*}{Can}
& Generated keyframe
& 45.0 & 0.0560
& 66.0 & 0.0578
& \textbf{81.0} & \textbf{0.0534} \\
& Real keyframe
& 67.0 & 0.0653
& 34.0 & 0.0598
& \textbf{79.0} & \textbf{0.0527} \\

\midrule

\multirow{2}{*}{Square}
& Generated keyframe
& 4.5 & 0.0335
& 13.5 & 0.0339
& \textbf{29.5} & \textbf{0.0333} \\
& Real keyframe
& 14.0 & 0.0332
& 10.5 & 0.0335
& \textbf{31.5} & \textbf{0.0331} \\

\bottomrule
\end{tabular}
}
\end{minipage}
\hfill
\begin{minipage}[t]{0.37\textwidth}
\centering
\vspace{0pt}
\includegraphics[width=\linewidth]{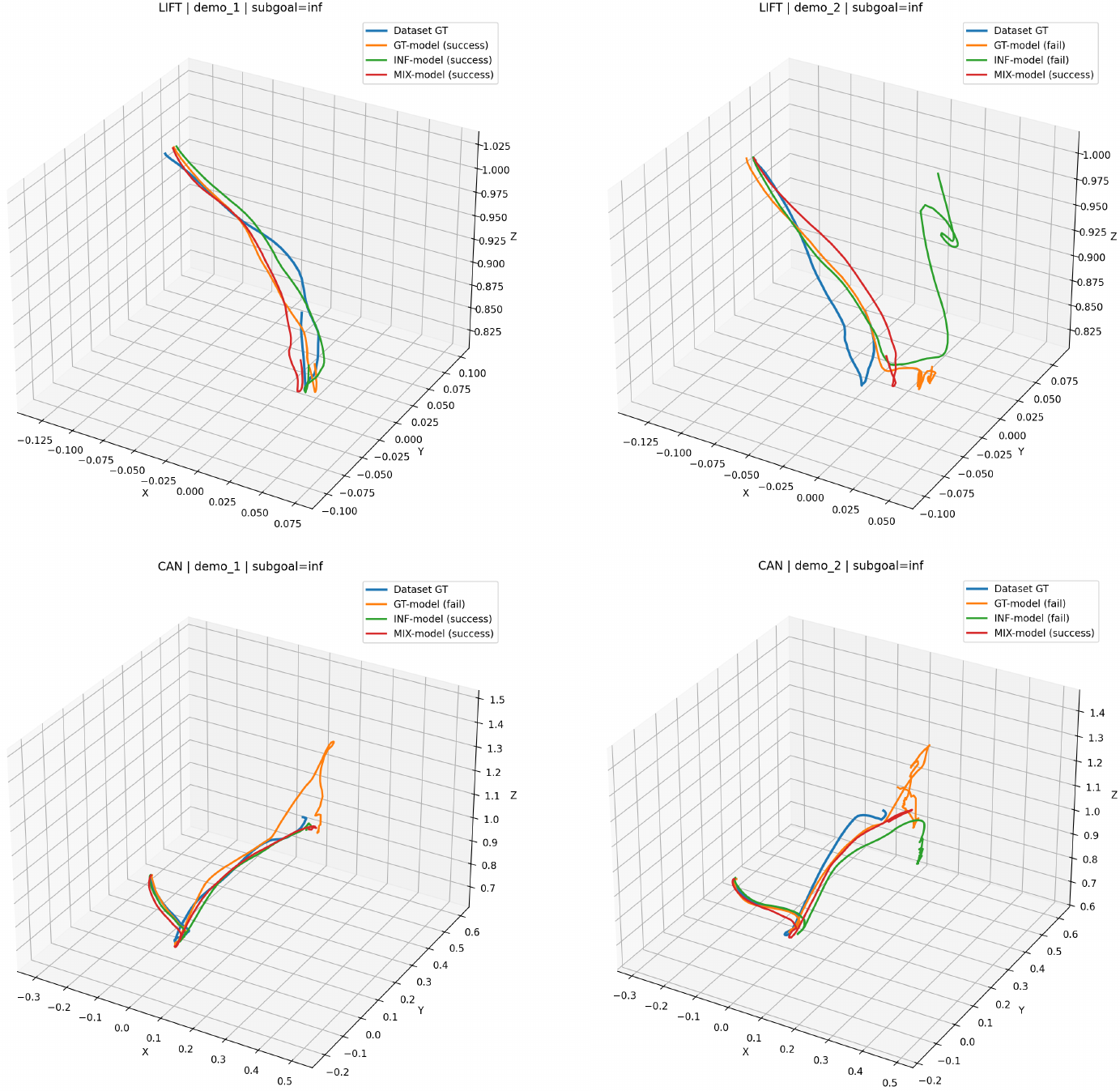}
\vspace{-3mm}
\captionof{figure}{Ablation visualization of action predictor trajectories.}
\label{fig:robot_execution_video}
\end{minipage}

\end{figure*}

\subsection{Full-Pipeline Execution on RoboMimic}
\vspace{-2mm}

We evaluate the full SWEET pipeline on the RoboMimic simulation benchmark to verify whether the generated keyframes can be translated into executable robot behaviors. 
As shown in Figure~\ref{fig:closed_loop_rollout}, the image-editing planner first predicts a sparse sequence of visual subgoals, and the goal-conditioned action predictor then executes action chunks between consecutive keyframes in a closed-loop manner. 
We test SWEET on three manipulation tasks, including Lift, Can, and Square, each trained with only a few hundred demonstrations. 
Table~\ref{tab:action_predictor_ablation_all} reports the task success rates, showing that SWEET can effectively close the loop from visual keyframe planning to robot control.

\subsection{Ablation Study}
\vspace{-2mm}

We study how the action predictor should be trained to execute planner-generated visual subgoals.
In this ablation, we keep the high-level planning results fixed and only vary the training data used for the action predictor.
Specifically, we compare three variants on RoboMimic: training with only real keyframes (\textit{Real-trained}), only generated keyframes from FLUX-Kontext (\textit{Gen-trained}), and a \(1{:}1\) mixture of both (\textit{Mix-trained}), while keeping the action labels unchanged.
For evaluation, we use the same planned keyframes as targets and report closed-loop task success rate and action prediction MSE.
The MSE is computed between the predicted gripper position at the ending timestep of each subtask and the corresponding ground-truth position.

As shown in Table~\ref{tab:action_predictor_ablation_all}, the action predictor is sensitive to the target-keyframe domain.
For example, on Lift, Real-trained performs well with real-keyframe targets but degrades substantially when tested with generated targets, while Gen-trained shows the opposite tendency.
In contrast, Mix-trained achieves consistently strong performance under both target types, indicating better robustness to the real-to-edited subgoal gap.
Similar trends are observed on Can and Square.
Although Square is more challenging because the robot must precisely insert the square nut onto the peg, leaving little tolerance for small gripper-trajectory errors, Mix-trained still outperforms the single-domain variants.
Overall, mixed training improves closed-loop execution while usually reducing gripper-position error.
These results suggest that edited-transition augmentation helps bridge the mismatch between real and planner-generated subgoals, enabling more reliable execution of visual plans.

\section{Limitations and Future Work}

Our work has two main limitations. 
First, scaling up the keyframe dataset remains challenging. 
Although large-scale robot video datasets are increasingly available, they usually do not provide the task-critical keyframe annotations required by our framework. 
A natural solution is to automatically extract keyframes from videos, but we found this still unreliable for manipulation tasks. 
While VLMs such as Gemini-3 can annotate arrow cues reasonably well when given sufficient context, they often fail to precisely identify contact or state-transition frames in robot videos, sometimes selecting frames that are temporally misaligned by several steps. 
Therefore, to ensure data quality, we rely on human keyframe annotation in this work, which limits scalability. 
A promising future direction is to fine-tune video-language models on robot manipulation videos to improve contact-point and state-transition detection.

Second, the action predictor still has room for improvement. 
In some cases, the edited keyframes are visually close to the ground truth, but the predicted actions are still not accurate enough for successful execution, especially in tasks requiring precise gripper motion. 
This is partly due to the limited amount of action-prediction training data. 
A future direction is to jointly train visual prediction and action prediction, similar in spirit to World Action Models, so that generated visual subgoals and executable actions can be better aligned.








\section{Conclusion}
\vspace{-2mm}
We presented \textbf{SWEET}, a keyframe-based visual planning framework that revisits whether dense video prediction is necessary for robotic manipulation. 
Through controlled comparisons on DROID and RoboMimic, we find that image editing models can serve as promising visual planners, producing task-relevant future keyframes with favorable visual quality, physical plausibility, and efficiency compared with video generation baselines. 
SWEET further connects these generated keyframes to executable control through a goal-conditioned diffusion action predictor. 
To reduce the domain gap between real and planner-generated subgoals, we introduce a direct mixed-training strategy with filtered edited transitions, improving robustness in closed-loop execution. 
While our experiments are still preliminary, the results suggest that sparse image transformation is a promising and potentially scalable alternative to dense video rollout, opening a new direction for embodied visual planning.

\small 
\bibliographystyle{plainnat} 
\bibliography{main}

\newpage
\appendix

\setcounter{page}{1}

\setcounter{section}{0}
\renewcommand{\thesection}{\Alph{section}}

\end{document}